\documentclass[journal]{IEEEtran}
\usepackage{amsmath,amsfonts}
\usepackage{algorithmic}
\usepackage{algorithm}
\usepackage{array}
\usepackage[caption=false,font=normalsize,labelfont=sf,textfont=sf]{subfig}
\usepackage{textcomp}
\usepackage{stfloats}
\usepackage{url}
\usepackage{verbatim}
\usepackage{graphicx}
\usepackage{cite}
\usepackage[pagebackref,breaklinks,colorlinks]{hyperref}
\usepackage{color,xcolor}
\usepackage{fancyhdr}

\usepackage{scalerel}
\usepackage{tikz}
\usetikzlibrary{svg.path}

\definecolor{orcidlogocol}{HTML}{A6CE39}
\tikzset{
  orcidlogo/.pic={
    \fill[orcidlogocol] svg{M256,128c0,70.7-57.3,128-128,128C57.3,256,0,198.7,0,128C0,57.3,57.3,0,128,0C198.7,0,256,57.3,256,128z};
    \fill[white] svg{M86.3,186.2H70.9V79.1h15.4v48.4V186.2z}
                 svg{M108.9,79.1h41.6c39.6,0,57,28.3,57,53.6c0,27.5-21.5,53.6-56.8,53.6h-41.8V79.1z M124.3,172.4h24.5c34.9,0,42.9-26.5,42.9-39.7c0-21.5-13.7-39.7-43.7-39.7h-23.7V172.4z}
                 svg{M88.7,56.8c0,5.5-4.5,10.1-10.1,10.1c-5.6,0-10.1-4.6-10.1-10.1c0-5.6,4.5-10.1,10.1-10.1C84.2,46.7,88.7,51.3,88.7,56.8z};
  }
}

\newcommand\orcidicon[1]{\href{https://orcid.org/#1}{\mbox{\scalerel*{
\begin{tikzpicture}[yscale=-1,transform shape]
\pic{orcidlogo};
\end{tikzpicture}
}{|}}}}

\begin{document}
\title{Customizable ROI-Based Deep Image Compression}

\author{Jian~Jin$^{\orcidicon{0000-0003-4250-1519}*}$,
Fanxin~Xia$^{*}$,
Feng Ding, Xinfeng~Zhang$^{\orcidicon{0000-0002-7517-3868}}$, Meiqin~Liu, \\
Yao Zhao$^{\orcidicon{0000-0002-8581-9554}}$,~\IEEEmembership{Fellow,~IEEE},
Weisi~Lin$^{\orcidicon{0000-0001-9866-1947}}$,~\IEEEmembership{Fellow,~IEEE},
Lili~Meng 

\thanks{Jian Jin and Weisi Lin are with the College of Computing and Data Science, Nanyang Technological University, 639798, Singapore. Jian Jin is also with the Center of China-Singapore International Joint Research Institute, 510555, China. (e-mail: jianj008@gmail.com; wslin@ntu.edu.sg.)}
\thanks{Fanxin Xia and Lili Meng are with the School of Information Science and Engineering, Shandong Normal University, Jinan, 250014, China. (e-mail: xiafx979@gmail.com; mengll\_83@hotmail.com.)}
\thanks{Feng Ding is with the School of Engineering Science, Simon Fraser University, V5A 1S6, Canada (e-mail: feng\_ding@sfu.ca.)}
\thanks{Xinfeng Zhang is with the School of Computer Science and Technology, University of Chinese Academy of Sciences, Beijing, 100049, China (e-mail: xfzhang@ucas.ac.cn).}
\thanks{Meiqin Liu and Yao Zhao are with the Institute of Information Science, Beijing Jiao Tong
 University, 100044, China. (e-mail: mqliu@bjtu.edu.cn; yzhao@bjtu.edu.cn.)}
\thanks{Corresponding author: Weisi Lin and Lili Meng}
\thanks{* Equal contributions}
}

\markboth{Journal of \LaTeX\ Class Files,~Vol.~14, No.~8, August~2021}%
{Shell \MakeLowercase{\textit{et al.}}: A Sample Article Using IEEEtran.cls for IEEE Journals}


\maketitle
\thispagestyle{fancy}


\fancyhead{} 

\cfoot{\copyright~2025 IEEE. Personal use of this material is permitted. However, permission to use this material for any other purposes must be obtained from the IEEE by sending an email to pubs-permissions@ieee.org.}

\renewcommand{\headrulewidth}{0mm}

\begin{abstract}

Region of Interest (ROI)-based image compression optimizes bit allocation by prioritizing ROI for higher-quality reconstruction. However, as the users (including human clients and downstream machine tasks) become more diverse, ROI-based image compression needs to be customizable to support various preferences. For example, different users may define distinct ROI or require different quality trade-offs between ROI and non-ROI. Existing ROI-based image compression schemes predefine the ROI, making it unchangeable, and lack effective mechanisms to balance reconstruction quality between ROI and non-ROI. This work proposes a paradigm for customizable ROI-based deep image compression. First, we develop a Text-controlled Mask Acquisition (TMA) module, which allows users to easily customize their ROI for compression by just inputting the corresponding semantic \emph{text}. It makes the encoder controlled by \emph{text}. Second, we design a Customizable Value Assign (CVA) mechanism, which masks the non-ROI with a changeable extent decided by users instead of a constant one to manage the reconstruction quality trade-off between ROI and non-ROI. Finally, we present a Latent Mask Attention (LMA) module, where the latent spatial prior of the mask and the latent Rate-Distortion Optimization (RDO) prior of the image are extracted and fused in the latent space, and further used to optimize the latent representation of the source image. Experimental results demonstrate that our proposed customizable ROI-based deep image compression paradigm effectively addresses the needs of customization for ROI definition and mask acquisition as well as the reconstruction quality trade-off management between the ROI and non-ROI. Additionally, even by using the uniform mask as input, our method still outperforms the anchor methods in image reconstruction and machine vision tasks (such as object detection and instance segmentation). Our source code will be available at: {\hypersetup{urlcolor=red}\url{https://github.com/hccavgcyv/Customizable-ROI-Based-Deep-Image-Compression}}

\end{abstract}


\section{Introduction}
\label{sec:intro}
\begin{figure}[!t]
\centering
\includegraphics[width=3.5in]{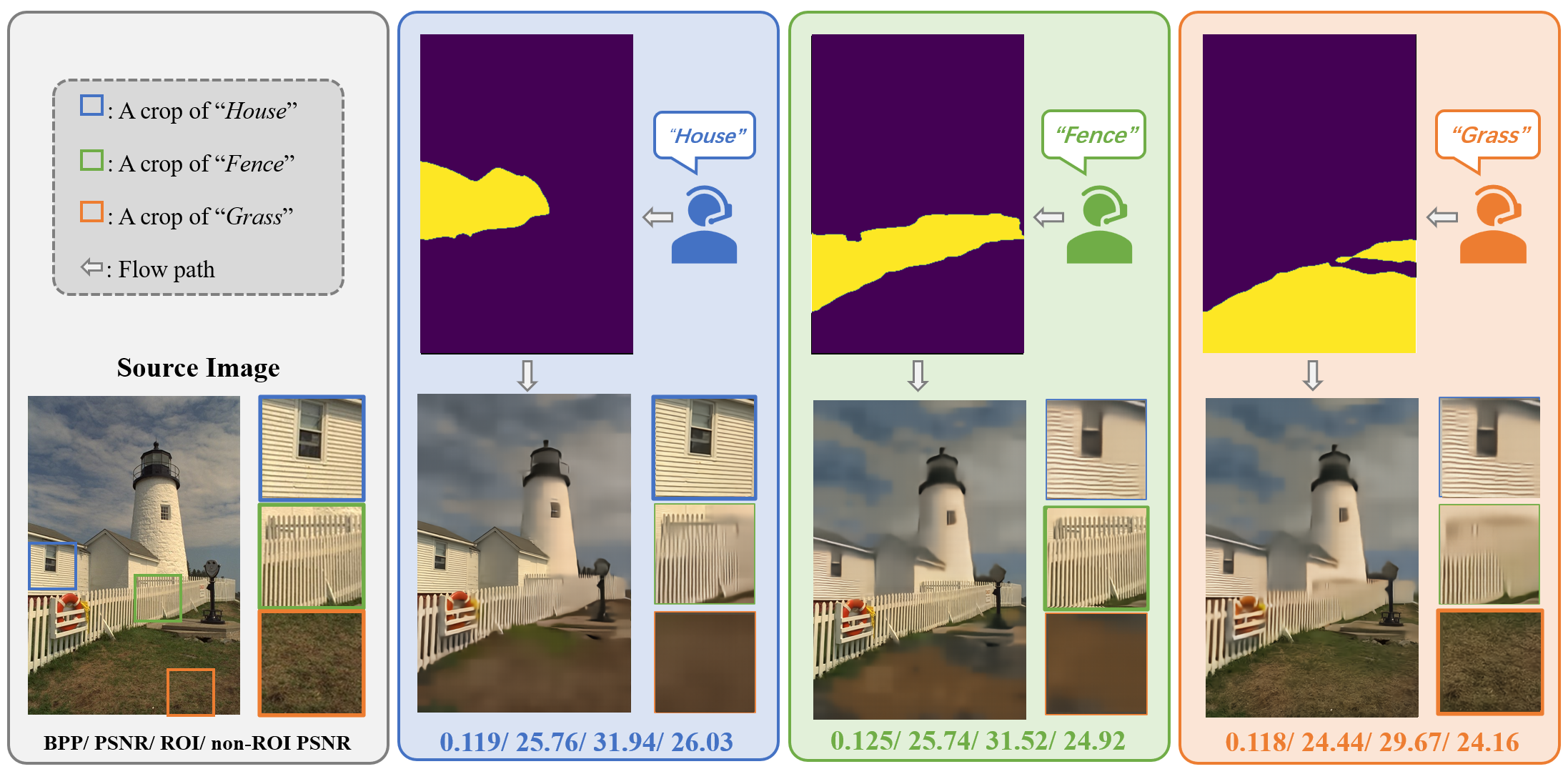}
\caption{The masks and their corresponding reconstructed images with our proposed paradigm by giving different semantic texts as inputs. Users customize their ROI with texts, \emph{i.e.}, \textcolor{blue}{\emph{``House''}}, \textcolor{green}{\emph{``Fence''}} and \textcolor{orange}{\emph{``Grass''}}, the corresponding masks can
be obtained, which are further used to perform ROI-based deep image compression. The enlarged crops demonstrate the effectiveness of the proposed \emph{customizable ROI-based deep image compression}. The test image is \emph{kodim19} from the Kodak \cite{kodak} dataset.}
\label{kodak19}
\end{figure}

\IEEEPARstart{G}{enerally}, the typical ROI-based image compression paradigm is performed in two phases: 1) acquiring the mask of ROI and non-ROI with the segmentation-relevant schemes, \emph{e.g.}, in video conferences \cite{borji2019salient, pang2020multi, wu2023hidanet}, video surveillance \cite{caelles2017one,zhang2020interactive,wu2024target}, and so on, and 2) encoding the masked ROI and non-ROI with high and low bitrates, respectively. Thus, the ROI-based image compression commonly preserves high reconstruction quality in the ROI, while achieving high compression efficiency in the non-ROI. So far, many efforts \cite{han2006image,li2023roi,kao2023transformer,lohdefink2020focussing,ma2021variable,xue2016fast,chen2021new} had been made to improve the compression efficiency of ROI while few studies had been done on customizable ROI-based image compression designs to cope with the various requirements of diverse users (including human clients and downstream machine tasks). As known, existing methods rely on Ground Truth (GT) masks or pre-acquired fixed masks (acquired with saliency detection \cite{10613841} or object segmentation methods \cite{tang2024semantic}) for compression. However, in practical applications, the definition of ROI and its mask acquisition should dynamically adapt to the preferences of specific users, as different users may pay attention to different regions even for the same image \cite{ding2023jnd}. For instance, in Fig. \ref{kodak19}, some users may focus on the house, while others may be interested in the fence or the grass. Similarly, for an image with the content of pedestrians and vehicles, the pedestrian detection task may care more about the pedestrian regions and select them as the ROI. In contrast, the vehicle identification task selects the vehicle regions as the ROI as it pays more attention to the vehicle regions. Therefore, \textbf{the definition of ROI and its mask acquisition should be customizable}. 

Besides, as aforementioned, the existing ROI-based image compression methods \cite{kao2023transformer,lohdefink2020focussing,chen2021new} mainly aim to preserve the reconstruction quality in ROI while neglecting the reconstruction quality of non-ROI. However, different application scenarios have different requirements for the reconstruction quality trade-offs between ROI and non-ROI. For instance, in road traffic video surveillance\cite{fu2021let}, compared with the car flow (\emph{i.e.}, ROI), the backgrounds such as roads and buildings (\emph{i.e.}, non-ROI) are not important in most situations, which can be reconstructed even with poor quality, but they can also be important when vehicle accidents happened (as they will be used as a reference for making the judgment of the accidents) and need to be reconstructed with tolerable quality. Therefore, \textbf{the quality trade-off between ROI and non-ROI should also be customizable}.

Additionally, the existing ROI-based image compression methods take the mask as the spatial prior in the image space. For instance, early ROI-based hybrid image compression methods \cite{bartrina2009jpeg2000,wang2007automatic,phadikar2010roi} directly utilized the spatial information provided by the mask as the spatial prior to adjust the wavelet transform coefficient to achieve bit allocation. While for the ROI-based deep image compression methods \cite{xia2020object,li2023roi,kao2023transformer,lohdefink2020focussing,ma2021variable,dul2019object,chen2021new}, some initial works \cite{dul2019object,xia2020object,ma2021variable} treated the mask as the outline information and used the down-sampled mask as the spatial prior in the image space to guide the latent representation of the source image (features extracted from the source image by deep encoder) in the latent space. Besides, some works \cite{li2023roi,kao2023transformer,lohdefink2020focussing} tried to learn the latent representation of the source image by feeding it together with the mask as input and the mask was still used as the spatial prior in the image space. However, directly utilizing such spatial prior in the image space to guide the latent representation of the source image in the latent space is obviously coarse, as such spatial prior and latent representation exist in different spaces. Hence, ROI-based image compression requires finer spatial prior in the latent space (termed \emph{latent spatial prior}) for guidance. Besides, as no Rate-Distortion Optimization (RDO) \cite{shannon1948mathematical} prior is introduced to tell the network what details of latent representation contribute more to the source image reconstruction under the fixed bitrate, the latent representation of the source image is still learned from scratch. Therefore, to help with the RDO of the latent representation, RDO prior in the latent space (termed \emph{latent RDO prior}) is also necessary.   

In this paper, we propose a \emph{customizable ROI-based deep image compression} paradigm to address the issues above. To achieve flexible mask definition and acquisition based on the requirements of different users, we propose a customizable mask acquisition method, \emph{i.e.}, a Text-controlled Mask Acquisition (TMA) module, which contains two sub-modules: Similarity Generation (SG) and Adjustable Binarization (AB). Users just input the semantic \emph{text} of the regions that they care about to mask into the SG sub-module, and it can find each pixel of the image with a similarity to the semantic \emph{text}. After a similarity binarization operation in the AB sub-module, a mask related to the semantic \emph{text} is acquired. Therefore, the ROI can be defined and its mask can be acquired by just giving the semantic \emph{text} as input, which is more flexible. Moreover, the proposed SG sub-module is a variant of CLIP \cite{radford2021learning}, which also inherits the zero-shot capability of CLIP for the new categories. This also improves the generalization of the TMA for mask acquisition. 

Additionally, to better manage the reconstruction quality trade-off between ROI and non-ROI, we propose a Customizable Value Assign (CVA) mechanism for the similarity binarization of the non-ROI in the AB sub-module, where the ROI is masked with 1 while the non-ROI is masked with a changeable extent. Specifically, we define a quality trade-off factor $\sigma$ with the range of $[0,1]$. A larger value is assigned to $\sigma$ when the non-ROI is required to be reconstructed with better quality, or vice versa. Then, the reconstruction quality trade-off between ROI and non-ROI can be customized by users by assigning a value from $[0,1]$ to $\sigma$. 

Besides, to obtain the latent spatial prior for guiding the latent representation of the source image in the same space, we propose a Latent Mask Attention (LMA) module, which mainly contains two sub-modules, \emph{i.e.}, Mask Representation (MR) and Importance map Generation (IG) \cite{fu2023asymmetric}. With the proposed MR sub-module, the latent spatial prior is extracted from the mask. Besides, to avoid the latent representation of the source image learning from scratch, we involve the IG sub-module for ROI-based deep image compression, which is used to generate the importance maps, used as the latent RDO prior. The latent RDO prior suggests what details of latent representation contribute more to the high-quality image reconstruction, and this effectively improves the compression efficiency of the proposed paradigm. Then, such two kinds of prior are fused and used for guiding the latent representation of the source image.

In summary, our main contributions are listed as follows.
\begin{itemize}
    \item We propose a \emph{customizable ROI-based deep image compression} paradigm that addresses the different requirements of various users. With the proposed paradigm, the ROI is able to be customizable for compression and the reconstruction quality trade-off between the ROI and non-ROI is also customizable by users. All these designs facilitate user customization, enhance the user experience, and improve the applicability and functionality of ROI-based image compression. It gives ROI-based image compression a broader application prospect, especially for users with customization requirements.

    \item We propose a TMA module, which has multimodality interaction capability and quality customizable capability. On the one hand, with the TMA module, users easily define their ROI and acquire masks using their corresponding semantic \emph{text}. On the other hand, the TMA module uses a changeable extent instead of a constant one for the non-ROI mask via the proposed CVA mechanism, allowing users to precisely customize the reconstruction quality trade-off between ROI and non-ROI.

    \item We propose an LMA module, where the latent spatial prior is extracted from the mask for guiding the latent representation of the source image and achieves better Rate-Distortion (RD) performance, especially for the boundary regions of the ROI. Besides, to avoid the latent representation of the source image being learned from scratch, we also fuse the latent RDO prior with the latent spatial prior in the LMA, which further improves the RD performance of the entire image.  

\end{itemize}

In this paper, both traditional ROI-based hybrid image codecs and ROI-based deep image codecs have been extensively tested for image reconstruction and machine vision tasks at the COCO \cite{lin2015microsoft} and Kodak \cite{kodak} datasets, which is a new benchmark for ROI-based image compression. The experimental results demonstrate that the proposed paradigm can fully achieve the customization functions that we mentioned above. Besides, even by using the uniform mask as input, we can still achieve the SOTA performance not only in image reconstruction in terms of ROI-PSNR and average PSNR but also in machine vision tasks, such as object detection and instance segmentation in terms of mAP.  

\section{Related Work}
\label{sec:related work}

\subsection{Image Compression}
\label{sec:IC}
\noindent{\textbf{Deep Image Compression.}}
With the advancement of deep learning, many deep image methods \cite{balle2016end,balle2018variational,minnen2018joint,cheng2020learned,fu2021learned,lu2021transformer,fu2023asymmetric,johnston2018improved,li2018learning,koyuncu2024efficient,li2024webp,xia2023gan,jin2022auto} have emerged rapidly, attracting increasing attention. In 2016, Ballé \emph{et. al} \cite{balle2016end} proposed the first end-to-end deep image compression method, which was mainly made up of the convolutional autoencoder and the fully factorized entropy model. Subsequently, main efforts were made to refine the precision of the estimated entropy model through the hyperprior model \cite{balle2018variational}, context model \cite{minnen2018joint}, Gaussian Mixture Model (GMM) \cite{cheng2020learned} and Gaussian-Laplacian-Logistic Mixture Model (GLLMM) \cite{fu2021learned}, \emph{etc}. Besides, various transformations were also proposed to enhance the expression capability of the latent space, such as attention module \cite{cheng2020learned,koyuncu2024efficient}, Transformer block \cite{lu2021transformer,li2024webp}, \emph{etc}. Moreover, some methods \cite{fu2023asymmetric,johnston2018improved,li2018learning} specifically emphasized the issue of bitrate allocation, optimizing the distribution strategy of bitrates by introducing the importance map concept. 
The importance map is automatically optimized via the RDO loss function. Although the importance map has been proven to boost the performance of image compression, it's still not used for ROI-based deep image compression.   

\noindent{\textbf{ROI-Based Image Compression.}}
ROI-based image compression schemes acquire the mask with salient detection or object segmentation methods. For instance, Cai \emph{et al.} \cite{cai2019end} proposed an end-to-end ROI-based image compression scheme by using the salient detection method to obtain the mask of the face as ROI. Xia \emph{et al.} \cite{xia2020object} used a segmentation method in DeepLab\cite{chen2017deeplab} to obtain the mask of ROI and non-ROI and then encode them separately.
Besides, there are some works tied to using the masks in different ways to improve compression efficiency. For example, Li \emph{et. al} \cite{li2023roi} and Kao \emph{et. al} \cite{kao2023transformer} directly used the mask as prior information and concatenated it with the source image in the image space before feeding it into the network. Subsequently, after several downsampling layers, the Spatial Feature Transform (SFT) \cite{wang2018recovering} module or the Swin Transformer \cite{liu2021swin} module was used to further integrate the down-sampled mask with the latent representation of the source image. However, it's obviously not suitable to directly use the mask in the image space as the spatial prior for guiding the latent representation of the source image in the latent space, finer latent spatial prior is needed.

\subsection{Language-driven Visual Recognition}
Recently, language-driven visual recognition \cite{zhang2024language} has been an active research direction of multimodality \cite{jewitt2016introducing}. It aims to integrate the rich semantic information contained in language descriptions with visual data processing, enabling more sophisticated and context-aware visual recognition tasks, such as object detection \cite{wang2023detecting,du2022learning,zhou2021ecffnet}, object segmentation \cite{wu2022language,li2023open,babu2004video}, \emph{etc}. 
The CLIP model profoundly understands the relationship between textual descriptions and their corresponding image content, granting CLIP with strong zero-shot \cite{zhang2023atzsl} capability. 
Given this, many works utilized the CLIP to solve typical vision tasks. For instance, CLIPasso \cite{vinker2022clipasso} leveraged CLIP to achieve a semantically-aware object sketching task, which transformed images into line sketches in the style of Picasso. Meanwhile, ViLD \cite{gu2021open} utilized CLIP to develop an efficient object detection model. Similarly, Lseg \cite{li2022languagedriven} adopted CLIP's zero-shot ability to implement semantic segmentation tasks by inputting categories as prompts. Although many language-driven vision tasks benefit from it, little research has been conducted on ROI-based image compression.
\begin{figure*}[t]
\centering
\includegraphics[width=15cm]{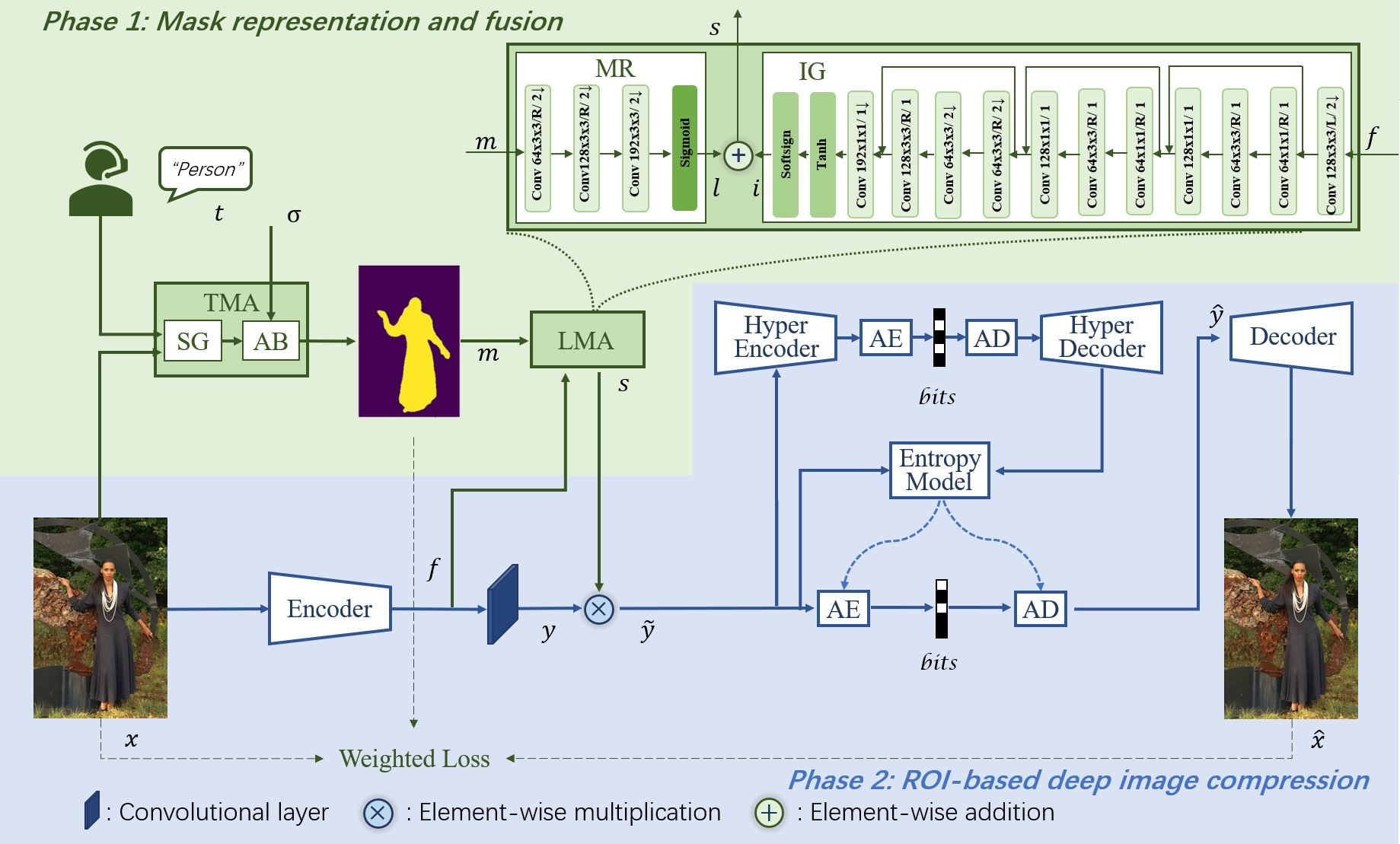}
\caption{Architecture of the proposed \emph{customizable ROI-based deep image compression} paradigm. It is achieved in two phases. Phase 1: mask representation and fusion process. Phase 2: ROI-based deep image compression process. 
}
\label{network}
\end{figure*}

\section{Method}

\subsection{Framework Overall}

The proposed \emph{customizable ROI-based deep image compression} paradigm is achieved by two phases, as shown in Fig. \ref{network}. Phase 1 is the mask representation and fusion process, including the proposed Text-controlled Mask Acquisition (TMA) and the Latent Mask Attention (LMA) modules. Our contributions mainly lie in phase 1.  

Phase 2 focuses on the compression part. Here, we adopt a typical deep image codec \cite{lu2021transformer} to conduct the image compression, which mainly consists of the encoder, the decoder, and the entropy model. Besides, it's also used as the baseline for comparisons in this paper.

The input of the proposed paradigm includes source image $x$, semantic text \emph{t}, and the quality trade-off factor $\sigma$ ($\sigma \in [0,1]$, managing the reconstruction quality trade-off between the ROI and non-ROI). Specifically, given source image $x$, users will define their ROI by inputting its associated semantic \emph{text} \emph{t} as the keyword. For instance, as shown in Fig. \ref{network}, the user defines the ROI with \emph{``Person''}. After that, both source image $x$ and the semantic text \emph{``Person''} are fed into the TMA module. Meanwhile, the quality trade-off factor $\sigma$ is also assigned by the user and fed into the TMA module. Then, a mask $m$ is obtained (the details are to be introduced in Sec. \ref{TMA}). It should be noted that if no text and value of $\sigma$ are input, the paradigm is performed with default settings, namely $t$ = \emph{``Foreground''}, and $\sigma$ = 0.01. In this case, the entire image will be defined as the ROI, which is like an ordinary deep image codec. Meanwhile, the source image $x$ is also fed into the encoder to extract the intermediate latent representation of source image $x$, termed $f$. On the one hand, the intermediate latent representation $f$ together with the mask $m$ are fed into the LMA to obtain the latent spatial prior $l$ and latent RDO prior $i$, which are then fused as the latent attention map $s$ (the details are to be introduced in Sec. \ref{LMA}). On the other hand, the intermediate latent representation $f$ passes through 1 convolutional layer and obtains the latent representation of the source image, termed $y$. Then, an element-wise multiplication operation is applied on $y$ and $s$. After that, an optimized latent representation of the source image is obtained, denoted by $\tilde{y}$. Then, $\tilde{y}$ is fed to the Hyper Encoder and Entropy Model to obtain the final bitstreams. After receiving the bitstreams, the image $\hat{x}$ is reconstructed on the decoder.

\subsection{Text-controlled Mask Acquisition}
\label{TMA}
We propose the TMA module to acquire the mask $m$ according to the user's requirements. On the one hand, the TMA module can acquire the mask of ROI that the user defined by just giving its corresponding semantic \emph{text} as input. On the other hand, the TMA module can manage the reconstruction quality tread-off between ROI and non-ROI by assigning the quality trade-off factor $\sigma$ with different values in $[0,1]$. The TMA module mainly contains the Similarity Generation (SG) and Adjustable Binarization (AB) sub-modules.

\noindent{\textbf{Similarity Generation.}}
\label{sec:VL}
To bridge the relationship between the text and image, we generate the similarity between the semantic \emph{text} and the corresponding regions of the image by absorbing the image encoder and text encoder components from the Lseg \cite{li2022languagedriven} (a variant of CLIP used for language-driven semantic segmentation) as the SG sub-module. Given the semantic \emph{text} $t$ and its associated image $x$, the similarity $p_{hw}$ of pixel $(h,w)$ to the semantic \emph{text} is then calculated in the SG.
Specifically, we embed the semantic \emph{text} $t$ into a continuous vector space $\mathbb{R}^{C}$ with the text encoder of the SG. 
Similarly, the image encoder is used to produce an embedding vector for each pixel in $x$, and then we calculate the similarity between the input semantic \emph{text} and the corresponding regions in the image. Assume $H \times W \times C $ is the size of input image $x$ and $s$ is a scaling factor. We define $ \tilde{H} = {H}/{s}$, $ \tilde{W} = {W}/{s}$. The output is a dense embedding $I \in \mathbb{R} ^ {\tilde{H} \times \tilde{W} \times C}$. Similar to the Lseg, we set $s$ to 2 for efficient computing in this work. 
We refer to the embedding of pixel $(h,w)$ as $I_{hw}$. The outputs of text and image encoders are then used to calculate the cosine similarity \cite{gomaa2013survey}, formulated as
\begin{equation}
p_{hw} = \frac{\mathbf{I_{hw}} \cdot \mathbf{T}}{\|\mathbf{I_{hw}}\| \|\mathbf{T}\|}
\end{equation}
The output $p$ indicates the similarity between the semantic \emph{text} and each pixel. The numerical range of $p$ is $[0,1]$.

\begin{figure}[!t]
\centering
\includegraphics[width=3.5in]{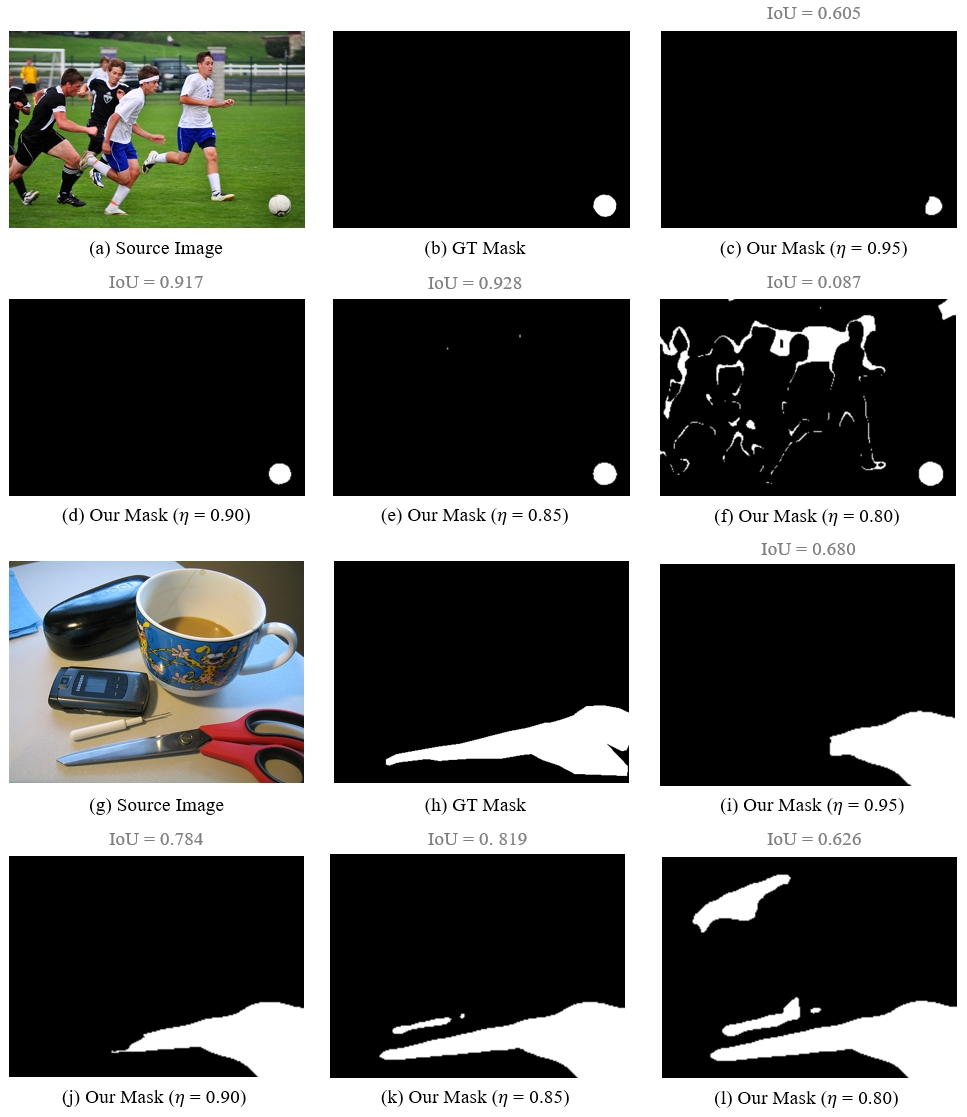}
\caption{Visualization of our masks obtained by assigning different values to $\eta$. (a) and (g) are the source images. Their GT segmentation masks of \emph{``sports ball''} and \emph{``scissors''} are (b) and (h). (c) to (f) and (i) to (l) are the masks obtained by inputting semantic texts with \emph{``sports ball''} and \emph{``scissors''} and assigning different values to $\eta$. To accurately evaluate the mask, we also calculate the IoU between $m$ and its associated GT segmentation mask in both simple single-target (a) and complex multiple-target (g) scenarios. It can be observed that a good result of $m$ is obtained in both of such two scenarios when $\eta$ is set to 0.85.}  
\label{mask}
\end{figure}

\noindent{\textbf{Adjustable Binarization.}}
\label{sec:AB}
To obtain the mask $m$, the similarity $p$ is to be binarized. Commonly, ROI and non-ROI are binarized with one and zero. However, as aforementioned, different scenarios require different reconstruction quality trade-offs between ROI and non-ROI.
To accurately manage the reconstruction quality trade-off, the ROI is still assigned to one during binarization. For the non-ROI, we propose a CVA mechanism where a changeable extent is used instead of a constant one for masking the non-ROI. 
Specifically, we first define a quality trade-off factor $\sigma$ $\in$ [0, 1].
Then, users can experientially decide the exact values from $[0,1]$ for $\sigma$ to balance the reconstruction quality between ROI and non-ROI. Generally, a larger value of $\sigma$ (close to 1) is assigned when a high-quality reconstruction of the non-ROI (comparable quality with that of ROI) is required. By contrast, a smaller value of $\sigma$ (close to 0) is assigned when a low-quality reconstruction of the non-ROI is needed. 
Mathematically, given the similarity $p$, the binarized mask $m$ is obtained with the formula below,
\begin{equation}
\label{similarity}
    m = Q(p;1,\sigma),
\end{equation}
where
\begin{equation}
Q(p_{hw})=
\left\{
	\begin{array}{cl}
		1, & \text{if } \eta < p_{hw} \leq1 \\
		\sigma, & \text{if } 0\leq p_{hw} \leq \eta, \sigma = 0,0.1,0.2, \ldots,1
	\end{array}.
\right.
\end{equation}
$\eta$ is an empirical value, which is used to control the mask of ROI. As $\eta$ gets smaller, more regions will be masked as the ROI. For instance, in Fig. \ref{mask}, we assign different values to $\eta$ and then visualize $m$ after upsampling and binarizing. It can be observed that when $\eta$ is set to 0.85, we obtain comparative results of the mask $m$ compared with the GT of the segmentation mask in both simple single-target and complex multiple-target scenarios. In this paper, we set $\eta$ to 0.85 after conducting lots of experiments. More details on the impacts of $\eta$ setting refer to Sec. \ref{Ablation}. 

Then, the latent spatial prior $l$ is extracted from the binarized mask $m$ in the LMA module, as to be introduced in the next part. Such a kind of prior will be used to guide the latent representation $y$ of the source image $x$, especially for managing the reconstruction quality of non-ROI. Moreover, the larger value of $\sigma$ indicates more information retained and hence better reconstruction quality of the non-ROI. By contrast, the smaller the value of $\sigma$, the less information is retained and the lower the reconstruction quality of the non-ROI. More results and discussions refer to Sec. \ref{sec_sigma}. 

\noindent{\textbf{Weighted Distortion Loss.}}
As aforementioned, by utilizing the CVA mechanism in the AB sub-module, users can control the reconstruction quality of non-ROI and further manage the reconstruction quality trade-off between ROI and non-ROI to some degree.
To strengthen such quality management ability, we also introduce the mask $m$ into the weighted distortion loss. Specifically, we first scale the size of mask $m$ from $\tilde{H} \times \tilde{W} \times 1$ to $H \times W \times 3$ to match the size of the source image. 
This is achieved by channel-wise repeting, followed by 
applying bilinear interpolation upsampling along the width and height dimensions. 
This process is represented with the function $U(\cdot)$. Then, we integrate it into the weighted distortion loss $D$ and we have 
\begin{equation}
\label{loss}
D = \Vert (x - \hat{x})\otimes U(m) \Vert_2.
\end{equation}
$\hat{x}$ denotes the reconstructed image of source image $x$. $\otimes$ is the element-wise multiplication. By assigning different values of $\sigma$ in $m$, users can effectively manage the reconstruction quality trade-off between ROI and non-ROI in $\hat{x}$.

\subsection{Latent Mask Attention (LMA)}
\label{LMA}
As aforementioned, the mask $m$ is a binary image with a single channel, which only contains the outline information in the image space. It's not reasonable to directly use its information in the image space to guide the latent representation $y$ of the source image in the latent space, as they are not in the same space. Therefore, we propose the LMA module. On the one hand, the LMA module is learned to extract the latent spatial prior $l$ from the mask $m$ in the latent space, which is more suitable for guiding the latent representation $y$ of the source image in the latent space. On the other hand, to avoid the latent representation $y$ being learned from scratch, the latent RDO prior $i$ is also fused with the latent spatial prior $l$ in the LMA module to optimize the latent representation of the source image for compression purposes. The LMA module mainly contains the Mask Representation (MR) and Importance map Generation (IG) sub-modules.

\noindent{\textbf{Mask Representation.}}
\label{sec:MR}
The MR sub-module is designed to obtain the latent spatial prior $l$ from the mask $m$. It consists of three convolutional layers and a sigmoid activation function. More architecture details of the MR are shown in Fig. \ref{network}. The mask $m$ is fed into the MR sub-module and then represented as the latent spatial prior $l$. $l$ has the same size as the latent representation $y$ of the source image. 

\begin{figure}[!t]
\centering
\includegraphics[width=3.5in]{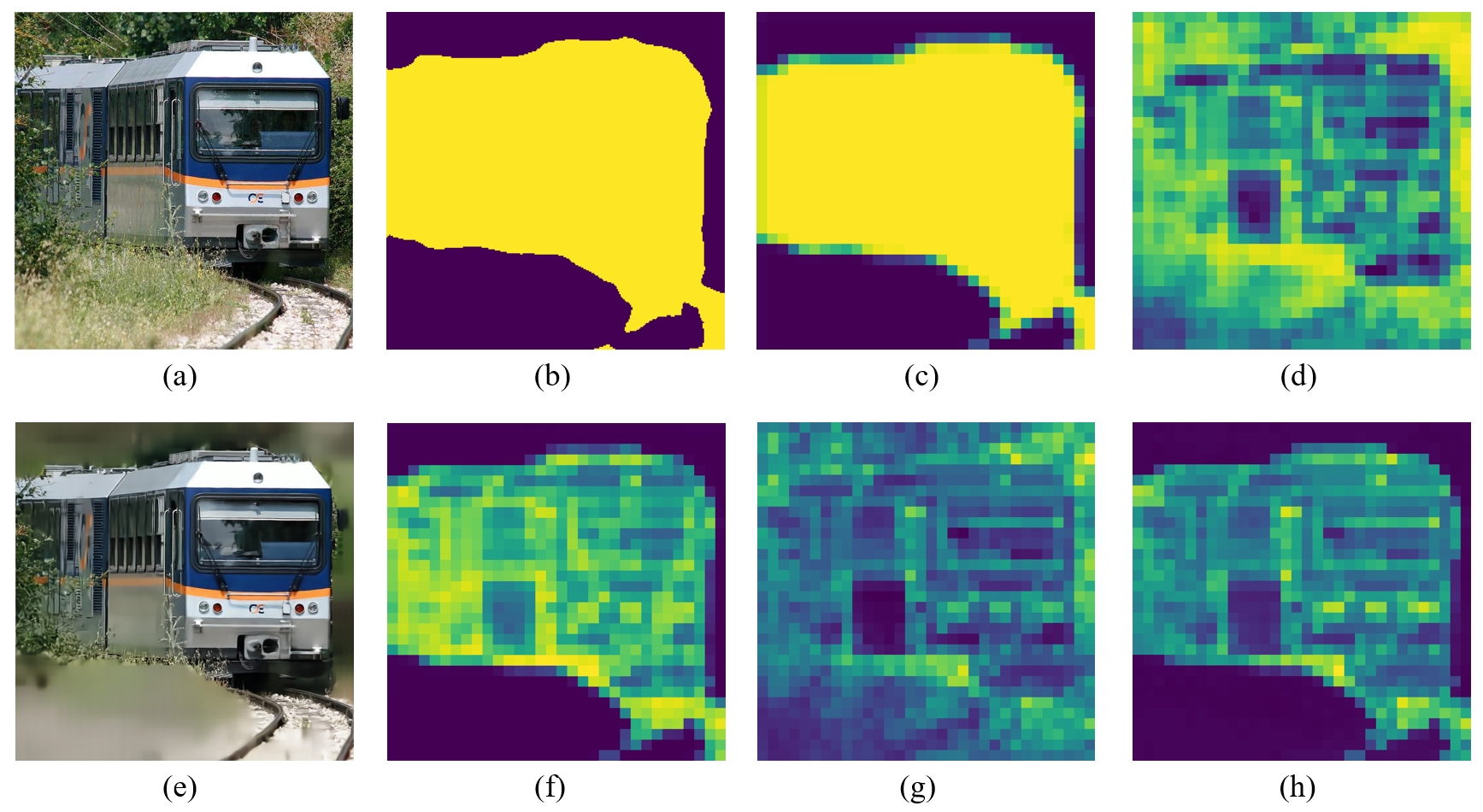}
\caption{Visualization of the variables in our paradigm. (a) and (b) are the source image $x$ and its associated mask $m$ when semantic text \emph{``train''} is given. (c) is the latent spatial prior $l$, extracted from the mask $m$. (d) is the associated latent RDO prior $i$. (e) is the reconstructed image $\hat{x}$. (f) is the attention map $s$. (g) and (h) are the latent representation $y$ and the optimized latent representation $\tilde{y}$ of the source image $x$, respectively.} 
\label{important_map}
\end{figure}

\noindent{\textbf{Importance Map Generation.}}
\label{sec:IG}
As aforementioned, the importance maps are used as the latent RDO prior in this paper which provides the information about which details of the latent representation $l$ contribute more to the reconstruction quality of the source image $x$. This is helpful for the bitrate allocation and improving the RD performance. 
To this end, we absorb the IG in \cite{fu2023asymmetric} to obtain the importance maps (\emph{i.e.}, latent RDO prior $i$) from the intermediate latent representation $f$ of the source image. 
The IG sub-module consists of two convolution layers, three residual blocks, and Tanh and Softsign activation functions, and the details are shown in Fig. \ref{network}. After feeding $f$ into the IG sub-module, we generate the latent RDO prior $i$. Similarly, $i$ has the same size as $l$. Then, we conduct the element-wise addition operation on $i$ and $l$ to obtain the fused feature $s$, which is defined as the attention maps in this paper. As the attention maps $s$ have fused the latent spatial prior and latent RDO prior, it helps guide the optimization of the latent representation of the source image in the following ROI-based image compression. 
More visualization results of the variables mentioned here refer to Fig. \ref{important_map} and a more detailed analysis based on such results will be introduced in Sec. \ref{Exp}.

\subsection{Implemention Details}
\noindent{\textbf{Dataset.}}
In this paper, three commonly used image datasets are used for training and testing, including Flicker 2W \cite{liu2020unified}, COCO \cite{lin2015microsoft}, and Kodak \cite{kodak}. For the training part, we first train the baseline (excluding Phase 1) on Flicker 2W, as there are no category annotations on it. After that, we train the whole paradigm (including Phase 1 and 2) on the COCO dataset by just using its images together with the category annotations. Besides, for the Flicker 2W dataset, we refer to its official preprocessing method and each image is randomly cropped into fixed patches at a size of $256\times256\times3$. For the COCO dataset, all images are re-scaled to a size of $256\times256\times3$. 
To verify the performance of the proposed paradigm, we mainly evaluate it against the anchor methods on the COCO validation dataset. To explore the zero-shot ability of the proposed paradigm, we also conduct evaluations on the Kodak dataset.

\noindent{\textbf{Training Parameters.}}
We select the Adam as the optimizer with a batch size of 8 and set the learning rate to 0.001. The loss function is $\mathcal{L}=\lambda D + R$, where $D$ is the weighted distortion loss, as formulated in Eq. \eqref{loss}. 
$R$ is the sum of bitrates, more details refer to \cite{lu2021transformer}. $\lambda$ is a Lagrange multiplier to control the trade-off between the rate $R$ and distortion $D$. Here, we train the proposed paradigm with 5 different values of $\lambda$ ($\lambda \in \{0.0035, 0.013, 0.025, 0.0483, 0.0932\}$) to match different bitrates and obtain 5 models. We adopt 128 channels for the first 2 models corresponding to low-bitrates scenarios, while 192 channels are adopted for the remaining 3 models corresponding to high-bitrates scenarios.

\noindent{\textbf{Training Strategy.}}
As shown in Fig. \ref{network}, the entire architecture comprises six components, making global optimization through end-to-end training challenging.
Hence, we adopt a three-stage training strategy.
In stage one, as there is no pre-trained model of the baseline (including the encoder, the decoder, and the entropy model in Phase 2), we train it on the Flicker 2W by using the image data without any additional category annotations. In stage two, we load the well-trained model in stage one and focus on training the modules in Phase 1 on the COCO dataset. Specifically, we train the MR sub-module to learn the latent spatial prior from the mask while keeping the rest of the modules frozen. In stage three, the entire paradigm is fine-tuned based on the well-trained models above on the COCO dataset. As the TMA module requires text input, both the images and their associated category annotations on the COCO dataset are used for the training in the last two stages.

\begin{figure*}[!t]
\centering
\includegraphics[width=15cm]{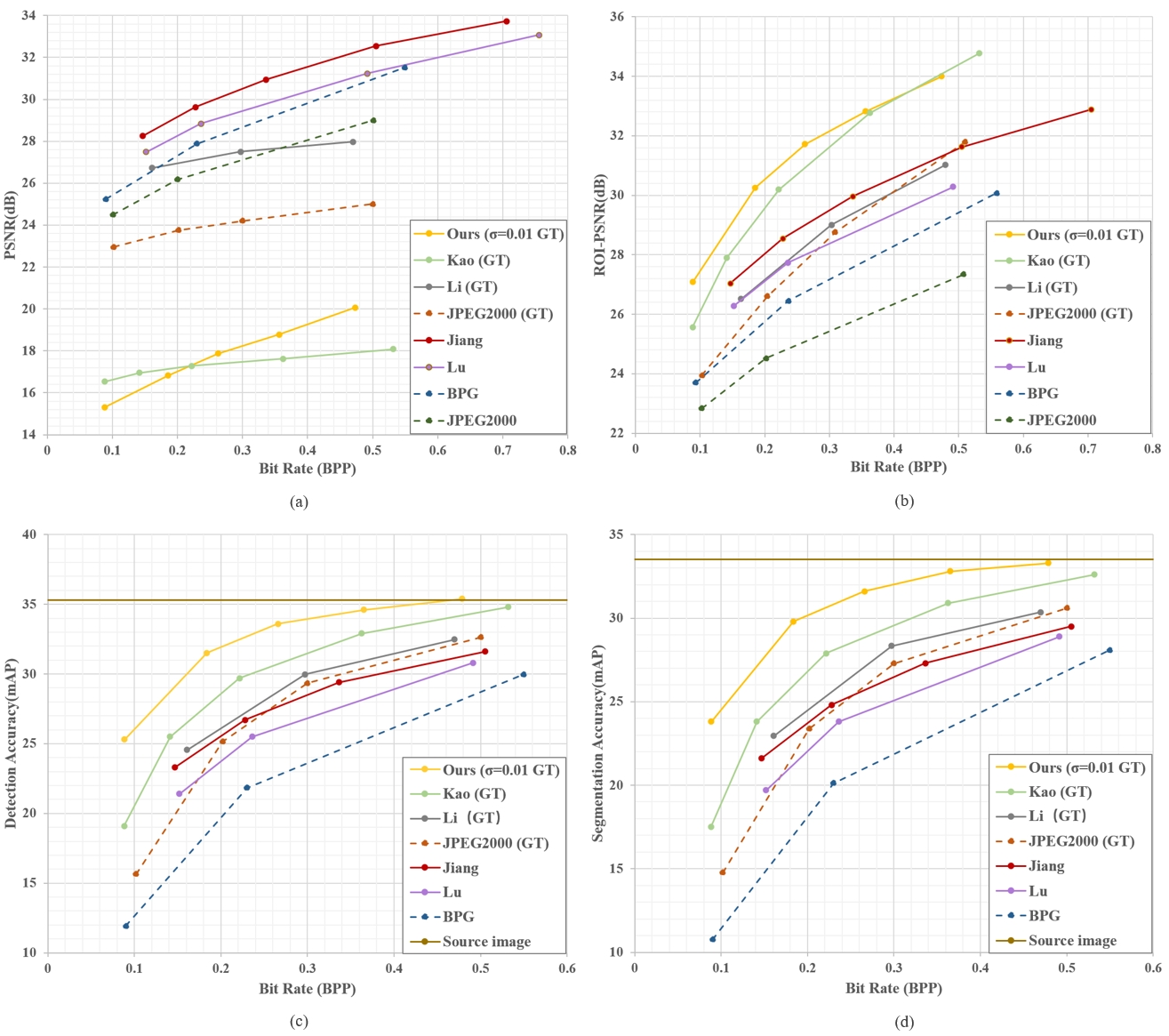}
\caption{RD performance on the COCO validation dataset. (a) is the reconstruction quality result of the entire image (PSNR \emph{vs} BPP). (b) is the reconstruction quality result of the ROI (ROI-PSNR \emph{vs} BPP). (c) and (d) are the object detection and instance segmentation results of the images compressed with different codecs (mAP \emph{vs} BPP), respectively. }
\label{RD}
\end{figure*}

\begin{figure*}[t]
\centering
\includegraphics[width=15cm]{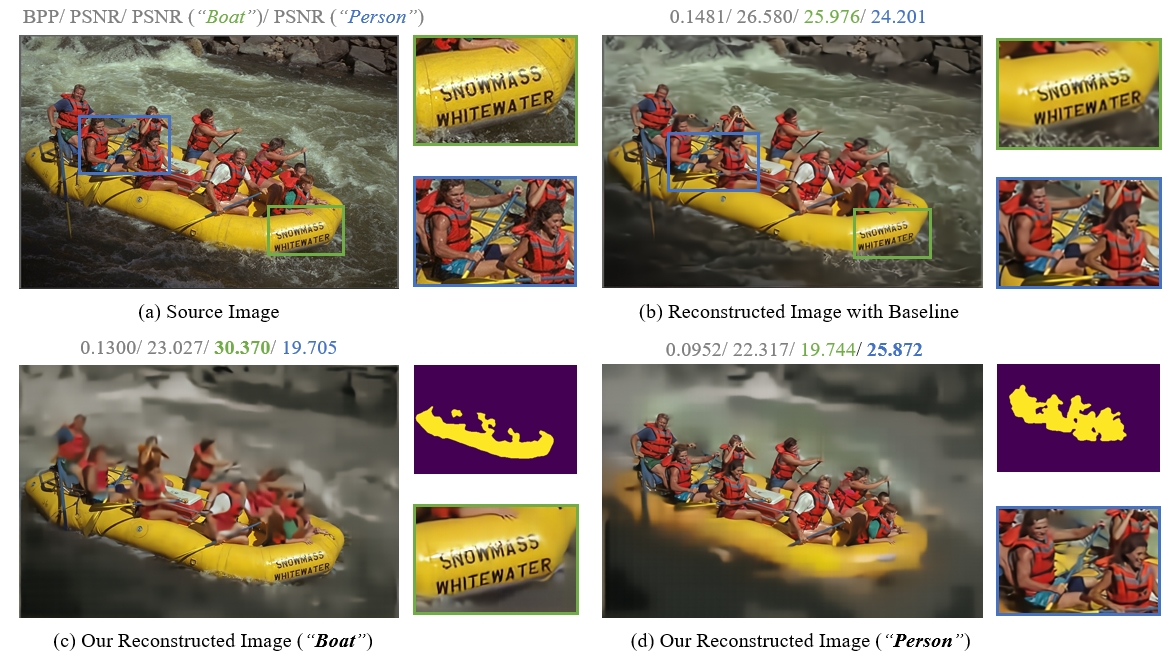}
\caption{The BPP, PSNR and ROI-PSNR comparisons among (a) the source image, (b) the reconstructed image obtained with the baseline \cite{lu2021transformer}, (c) the reconstructed image together with the mask obtained with our method by giving semantic text \emph{``Boat''}, and (d) the reconstructed image together with the mask obtained with our method by giving semantic text \emph{``Person''}.(c) and (d) are reconstructed with $\sigma$=0.1 and $\lambda$=0.0035.}
\label{kodak14}
\end{figure*}

\section{Experimental Results}
\label{Exp}
In this section, several typical ROI-based image compression methods are selected as anchors for comparisons, including Kao \cite{kao2023transformer} (the SOTA ROI-based deep image compression method), Li \cite{li2023roi}, JPEG2000 (ROI) \footnote{https://kakadusoftware.com}. Additionally, several other image compression methods (not for ROI-based image compression) are also selected for comparisons, including Jiang \cite{jiang2023mlicpp} (the SOTA deep image compression method), Lu\cite{lu2021transformer} (the baseline), JPEG2000 \cite{bartrina2009jpeg2000} and BPG \footnote{http://bellard.org/bpg/}.
To compare their performance, the reconstruction quality of the compressed images is evaluated with the metric Peak Signal-to-Noise Ratio (PSNR) of the entire image and PSNR of the ROI (ROI-PSNR) over various Bits Per Pixel (BPP), respectively. Except for the reconstruction quality being evaluated, two main machine tasks (object detection and instance segmentation) are verified on the reconstructed images with metrics mean Average Precision (mAP) over various BPP as well. Specifically, the Faster-RCNN \cite{ren2015faster} and Mask-RCNN \cite{he2017mask} networks with Resnet50 \cite{he2016deep} backbone \footnote{https://pytorch.org/vision/stable/models.html} are used for performing object detection and instance segmentation. It should be noted that the mAP results of the object detection and instance segmentation on source images are 35.3 $\%$ and 33.5$\%$, respectively.

\subsection{Rate-Distortion Performance Evaluation}

As reviewed before, the existing ROI-based image compression methods (including the anchor methods) can only use the pre-acquired masks for ROI-based image compression. In contrast, our proposed method is able to define the ROI for compression by giving its corresponding semantic \emph{text}. 
In view of this, for a fair comparison, we first utilize the GT of the segmentation results provided by the COCO dataset as the \emph{uniform mask} in the anchor methods and our proposed one. To this end, we replace the $p$ in Eq. \eqref{similarity} with the GT segmentation results while conducting our method. 

\begin{figure}[t]
  \centering  
  \includegraphics[width=3.2in]{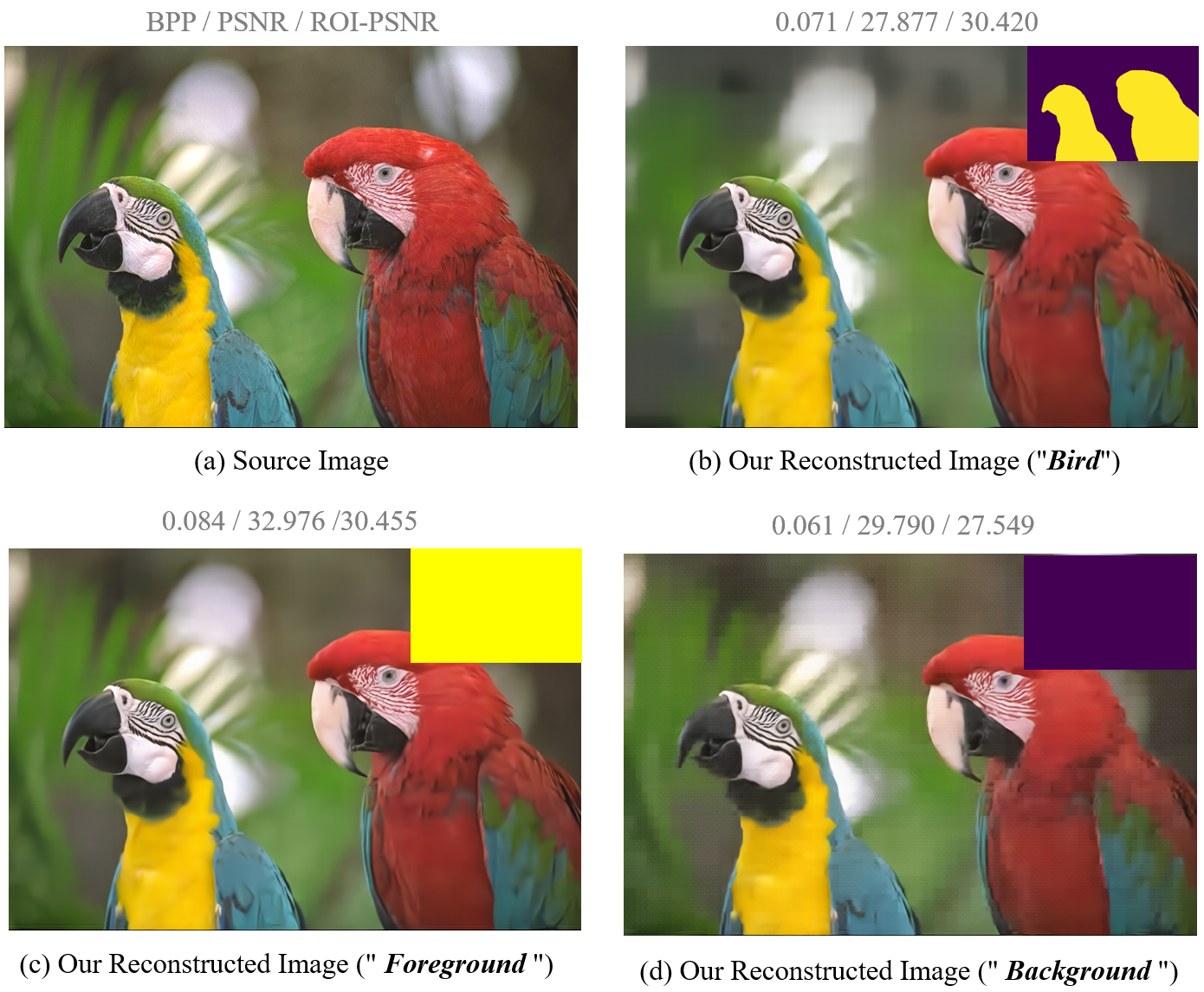}
   \caption{Our results on two special cases, \emph{i.e.}, the entire image is masked as the ROI and non-ROI, respectively. (a) is the source image. (b), (c) and (d) are our reconstructed images with different semantic texts \emph{``Bird''} (only the regions related to \emph{``Bird''} are masked as ROI), \emph{``Foreground''} (the entire image is masked as the ROI) and \emph{``Background''} (the entire image is masked as the non-ROI), respectively.
   All the iamge are reconstructed with $\sigma = 0.1$. (b) and (c) are reconstructed with $\lambda$ = 0.0035, while (d) are reconstructed with $\lambda$ = 0.0932.
   }
   \label{kodak_bird}
\end{figure}

\noindent{\textbf{PSNR and ROI-PSNR Results.}}
We first show the reconstruction quality results of the COCO validation dataset among the anchor methods and ours (with $\sigma=0.01$) in terms of the PSNR and ROI-PSNR in Fig. \ref{RD} (a) and (b), respectively. 
It can be observed that our method outperforms the anchor ones in ROI-PSNR, demonstrating its high compression effectiveness on ROI. But this comes at the cost of sacrificing the PSNR of the entire image. Although Kao's method has a similar ROI-PSNR to ours in the cases of high BPP, its PSNR of the entire image is smaller than ours. This also demonstrates the high compression effectiveness of our method.
Nevertheless, our approach can control the reconstruction quality of non-ROI and further manage the reconstruction quality trade-off between ROI and non-ROI by assigning different values to $\sigma$. More results and discussions are detailed in Sec. \ref{sec_sigma}.

\noindent{\textbf{Detection and Segmentation Accuracy Results.}}
We then show the object detection and instance segmentation results of the COCO validation dataset in terms of the mAP in Fig. \ref{RD} (c) and (d). 
It can be observed that our method achieved the best performance in both object detection and instance segmentation tasks under the same BPP.
Besides, our object detection result is nearly on par with the source image when BPP reaches 0.48. Also, Kao's method has a similar reconstruction quality of ROI to ours when BPP comes to 0.4, while our mAP is always larger than that of Kao's method.
The reason that our method can exceed Kao's method and even for the source image in object detection can be concluded as two benefits of our proposed method: 1) our method is able to better keep the information of ROI object while compressing the information in non-ROI to avoid non-ROI interference in object detection, and 2) we learn the latent spatial prior from the mask for ROI compression instead of the mask itself used in Kao's method, which makes the compressed images preserve finer information of the ROI object, especially for the object boundaries. Less interference information from non-ROI and finer boundary information from ROI make our method achieve SOTA results in object detection.

\subsection{Zero-shot Capability Evaluation}
Fig. \ref{kodak19} and Fig. \ref{kodak14} exhibit the results of our compressed images from the Kodak dataset. 
These results show that for the datasets without GT segmentation masks, our scheme can still achieve accurate ROI masking by simply providing a semantic \emph{text}, which significantly improves the applicability and functionality of the ROI-based image compression and it is also convenient for users.
Moreover, the semantic \emph{text} used in Fig. \ref{kodak19}, such as \emph{``house''}, \emph{``grass''}, and \emph{``fence''}, are not included in the categories of the COCO dataset. However, when using these keywords for mask acquisition, our scheme can still accurately identify the relevant regions. That is because the proposed SG sub-module is a variant of CLIP, which also inherits the zero-shot capability of CLIP. This also demonstrates that our method can help users achieve customizable ROI-based deep image compression by providing its corresponding semantic \emph{text}, even if the semantic \emph{text} does not appear in our training data.

\begin{figure}[t]
\centering
\includegraphics[width=2.0in]{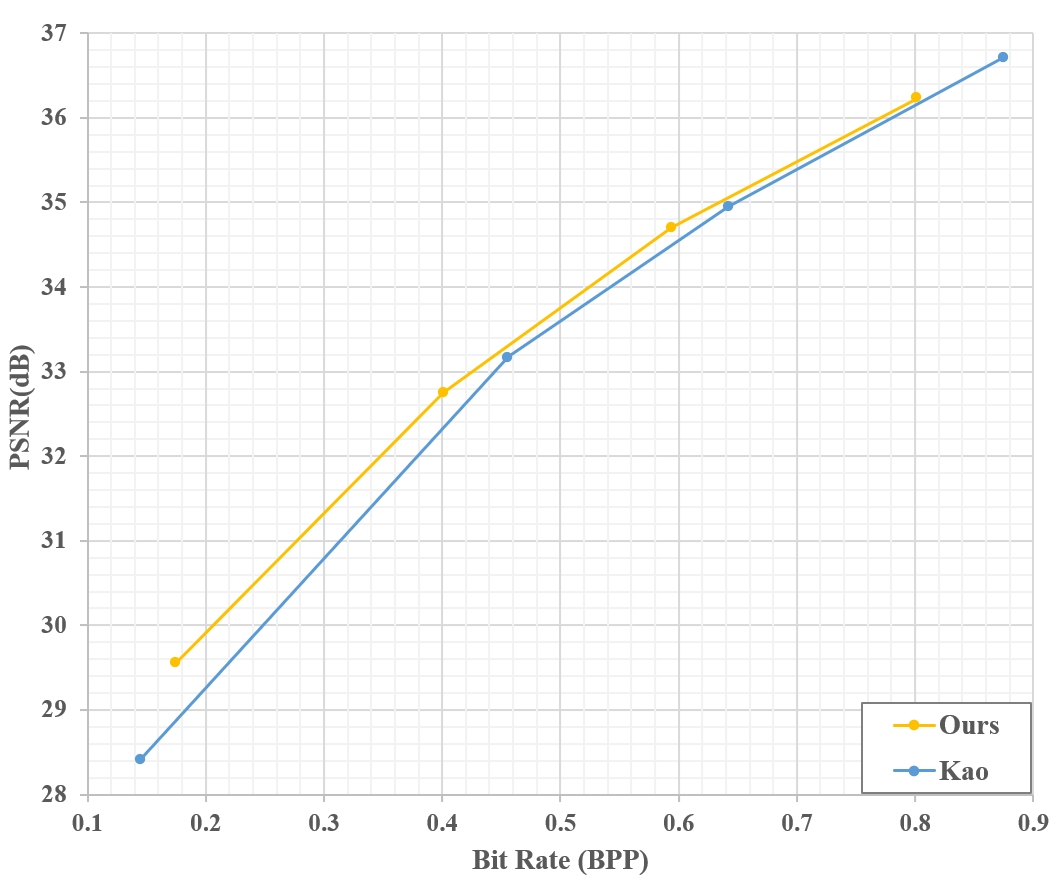}
\caption{RD performance comparisons by setting the entire image as the ROI in both Kao \cite{kao2023transformer} and our proposed method on the Kodak dataset.}
\label{kodak_roi}
\end{figure}

\begin{figure}[t]
\centering
\includegraphics[width=2.0in]{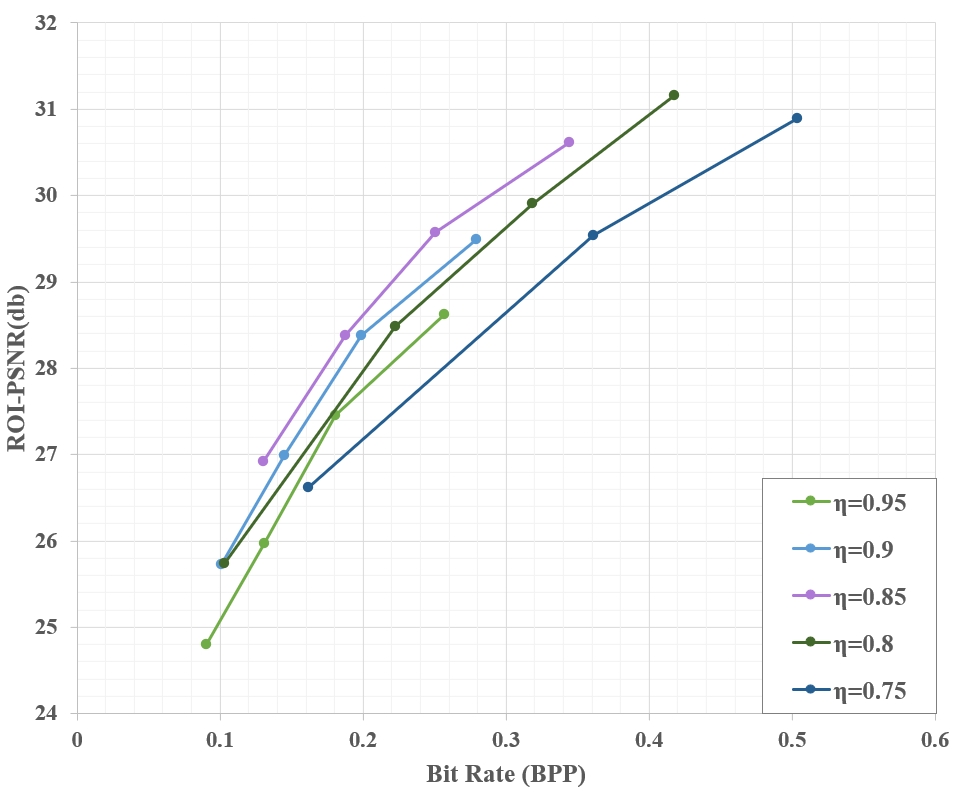}
\caption{RD performance comparisons with different $\eta$ settings on the COCO validation dataset. }
\label{eta_RD}
\end{figure}

\subsection{Special Cases Evaluation}

In this subsection, we first test the feasibility of the proposed method in two special cases: 1) the entire image is masked as the ROI (\emph{i.e.}, setting the value of $m$ to 1 when semantic \emph{text} is \emph{``Foreground''}, which is also the default setting when no semantic \emph{text} is given) and 2) the entire image is masked as the non-ROI (\emph{i.e.}, setting the value of $m$ to 0 when semantic \emph{text} is \emph{``Background''}), as shown in Fig. \ref{kodak_bird}. It can be observed that our method is still feasible even for these two cases. To further evaluate the performance of our proposed method in such two cases quantificationally, we select the SOTA ROI-based deep image compression method Kao as the anchor for comparison and evaluate both Kao's method and our proposed one on the Kodak dataset. The reconstruction quality against rate cost (PSNR \emph{vs} BPP) is shown in Fig. \ref{kodak_roi}. It can be observed that our proposed method outperforms Kao's one. The main reason is that the latent RDO prior $i$ is inherited in our method which provides the latent RDO prior for guiding the optimization of latent representation and achieves better compression results. 

\begin{figure*}[t]
\centering
\includegraphics[width=14cm]{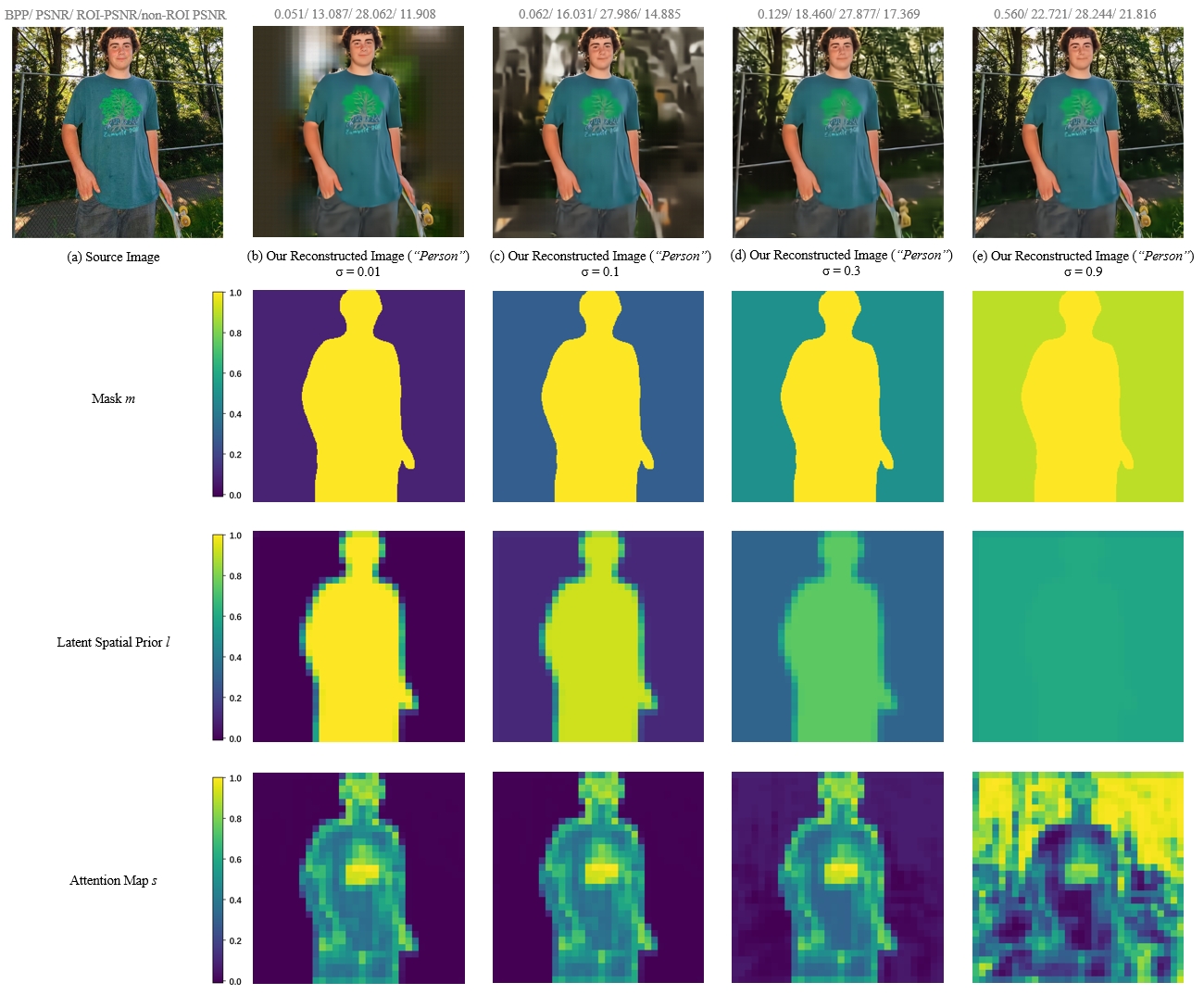}
\caption{Visualization of the reconstructed images, their associated masks, latent spatial prior, and attention maps with different settings of $\sigma$. }
\label{COCO_person}
\end{figure*}

\subsection{Ablation Studies}
\label{Ablation}
\noindent{\textbf{Setting of $\eta$.}}
The first group of ablation studies is conducted to justify the rationale for setting $\eta$ to 0.85 in this work. To this end, we test our method on the COCO validation dataset by assigning different values to $\eta$ ($\eta \in \{$0.95, 0.9, 0.85, 0.8, 0.75$\}$), and the results are shown in Fig. \ref{eta_RD}. 
The experimental results demonstrate our proposed method achieves the best result across all BPP points when $\eta$ is set to 0.85. Therefore, $\eta$ is set to 0.85 in this work.

\begin{figure}[t]
\centering
\includegraphics[width=3in]{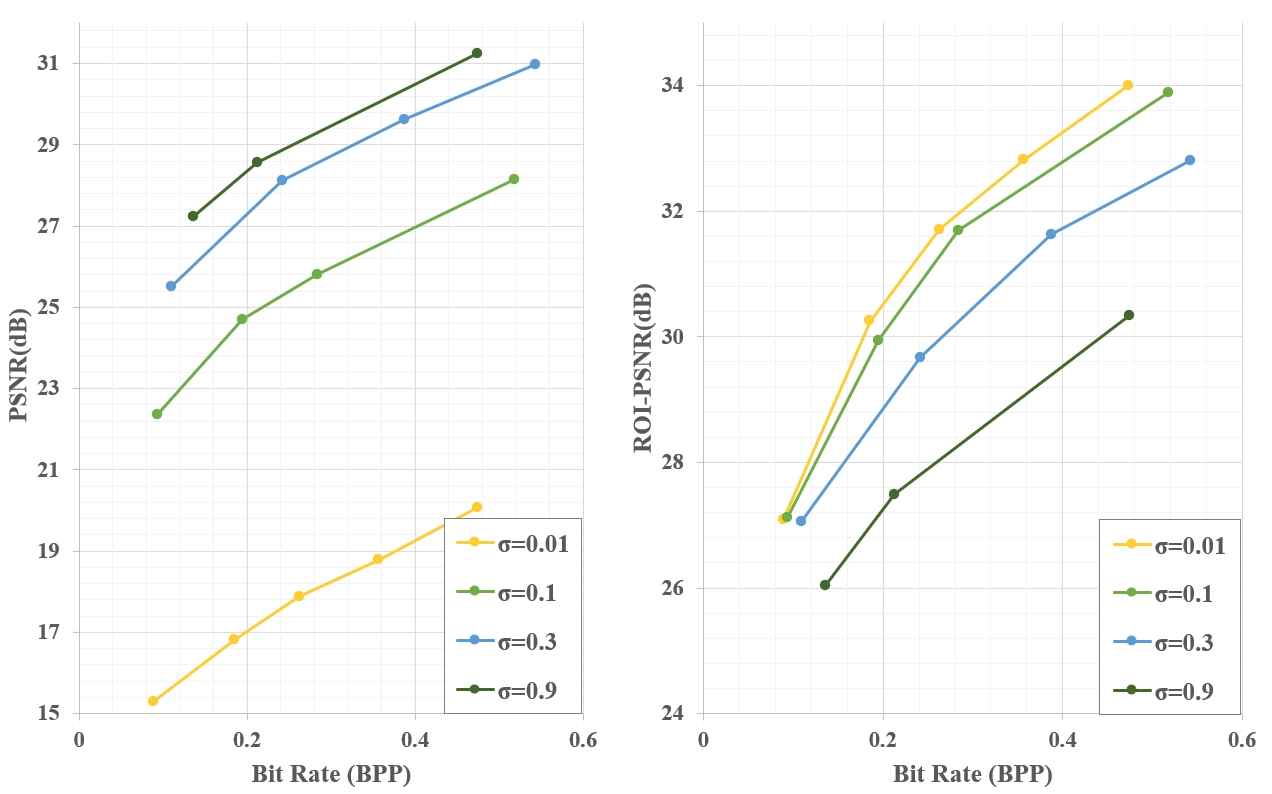}
\caption{RD performance comparisons with different $\sigma$ settings on the COCO validation dataset. }
\label{sigma}
\end{figure}

\noindent{\textbf{Setting of $\sigma$.}}
\label{sec_sigma}
The second group of ablation studies is conducted to validate the effectiveness of $\sigma$ in this work. As aforementioned, $\sigma$ is a factor in managing the reconstruction quality trade-off between the ROI and non-ROI. 
To validate its effectiveness, we visualize the results of the reconstruction images (in the first row), their associated mask $m$ (in the second row), and the latent spatial prior $l$ (in the third row) of the mask, as well as the attention map $s$ (in the last row) with different values of $\sigma$ ($\sigma \in \{$0.01, 0.1, 0.3, 0.9$\}$) in Fig. \ref{COCO_person}. As the $\sigma$ increases, the reconstruction quality of the non-ROI gradually improves and the mask values in ROI and non-ROI become closer. Similarly, their associated latent spatial prior have similar trends. Besides, the attention map $s$ in the last row visualizes the detailed bits allocation in latent space. It can be seen that as $\sigma$ increases, more bits are allocated in the non-ROI to achieve high-quality reconstruction of the non-ROI. 
Besides, we also test our proposed method on the entire COCO validation dataset with different $\sigma$ settings, and the results are shown in Fig. \ref{sigma}. It can be observed that as the value of $\sigma$ increases, the PSNR across all BPP points increases, while the ROI-PSNR decreases. This is because when $\sigma$ increases, non-ROI (with a high ratio of the entire image) are reconstructed with high quality, resulting in an increase in the PSNR of the entire image. Meanwhile, to balance the bit-rate allocation between the ROI and non-ROI, the reconstruction quality of the ROI decreases inevitably. 
This aligns with our claims on effectively managing the reconstruction quality trade-off between ROI and non-ROI.

\begin{figure*}[t]
\centering
\includegraphics[width=17.5cm]{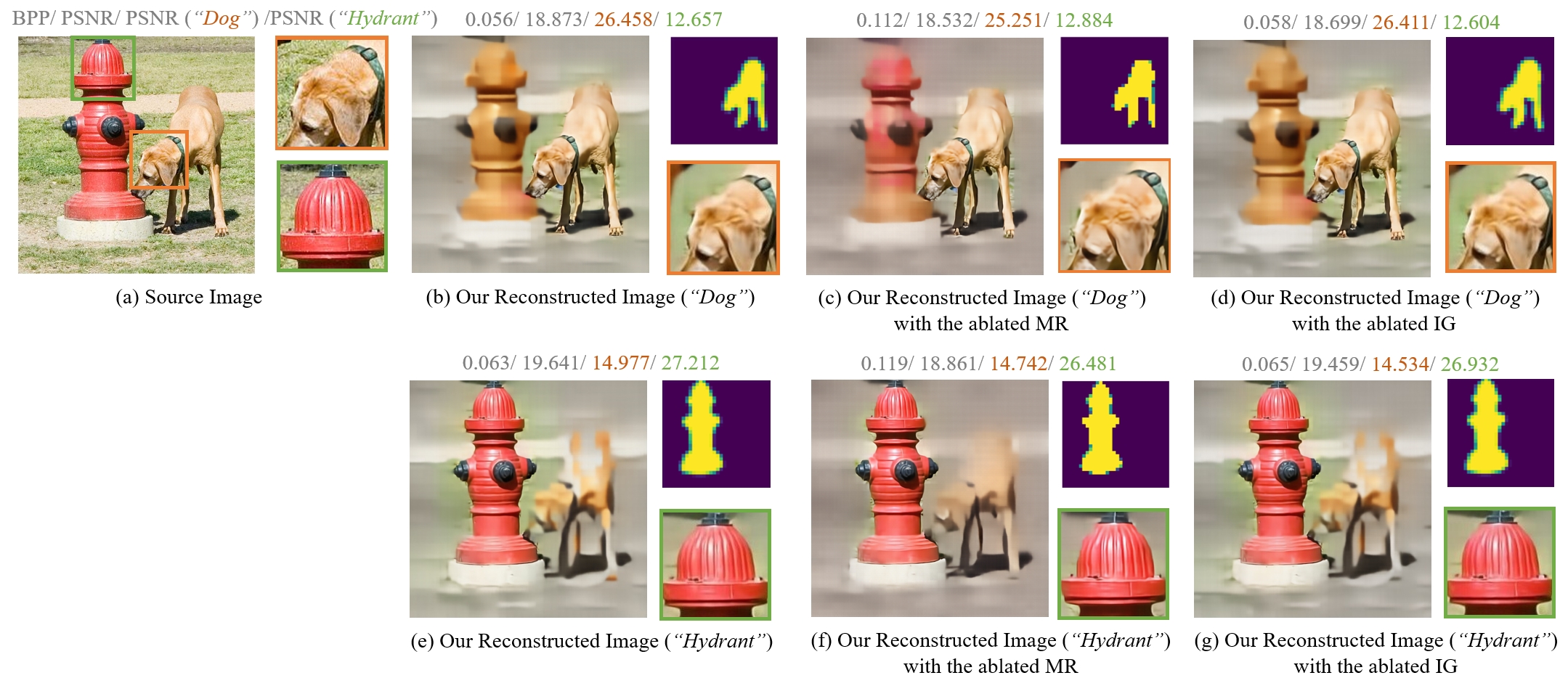}
\caption{Visualization result of the sub-module MR and IG ablation studies. (a) is the source image. (b) and (e) are our reconstructed images and the latent spatial prior of the masks obtained with our original MR and IG sub-modules by inputting semantic texts \emph{``Dog''} and \emph{``Hydrant''}, respectively. (c) and (f) are their corresponding results obtained with the MR ablation settings. Similarly, (d) and (g) are the results obtained with the IG ablation settings. All the image are reconstructed with $\sigma$=0.1 and $\lambda$=0.0035.}  

\label{COCO_dog}
\end{figure*}

\noindent{\textbf{MR ablation.}} To verify the effectiveness of the proposed MR sub-module, we replace the convolution operations of the MR sub-module with channel-wise repetition and bilinear interpolation downsampling operations. Specifically, the single-channel mask $m$ is first repeated in channel dimension to match the channel number of the latent representation $y$, followed by applying bilinear interpolation downsampling in width and height dimensions to be consistent with the size of $y$. The MR ablation results are shown in Fig. \ref{COCO_dog}, where the reconstructed images together with their associated latent spatial prior of the masks obtained with our original MR sub-module and the ablated MR sub-module are shown in the second column (\emph{i.e.}, (b) and (e)) and the third column (\emph{i.e.}, (c) and (f)) of Fig. \ref{COCO_dog}. It can be observed that the latent spatial prior of the masks obtained with the original MR sub-module in (b) and (e) have a better transition at the outline compared with MR ablation results in (c) and (f). Besides, both the subjective and objective (PSNR) reconstruction quality at the outline regions in (b) and (e) is better compared with that in (c) and (f) even under smaller BPP conditions. This demonstrates that our proposed MR sub-module is able to learn a finer latent spatial prior of the mask for better guiding the latent representation of the source image. This also aligns with our claims on the effectiveness of the latent spatial prior.

\begin{figure}[t]
\centering
\includegraphics[width=3in]{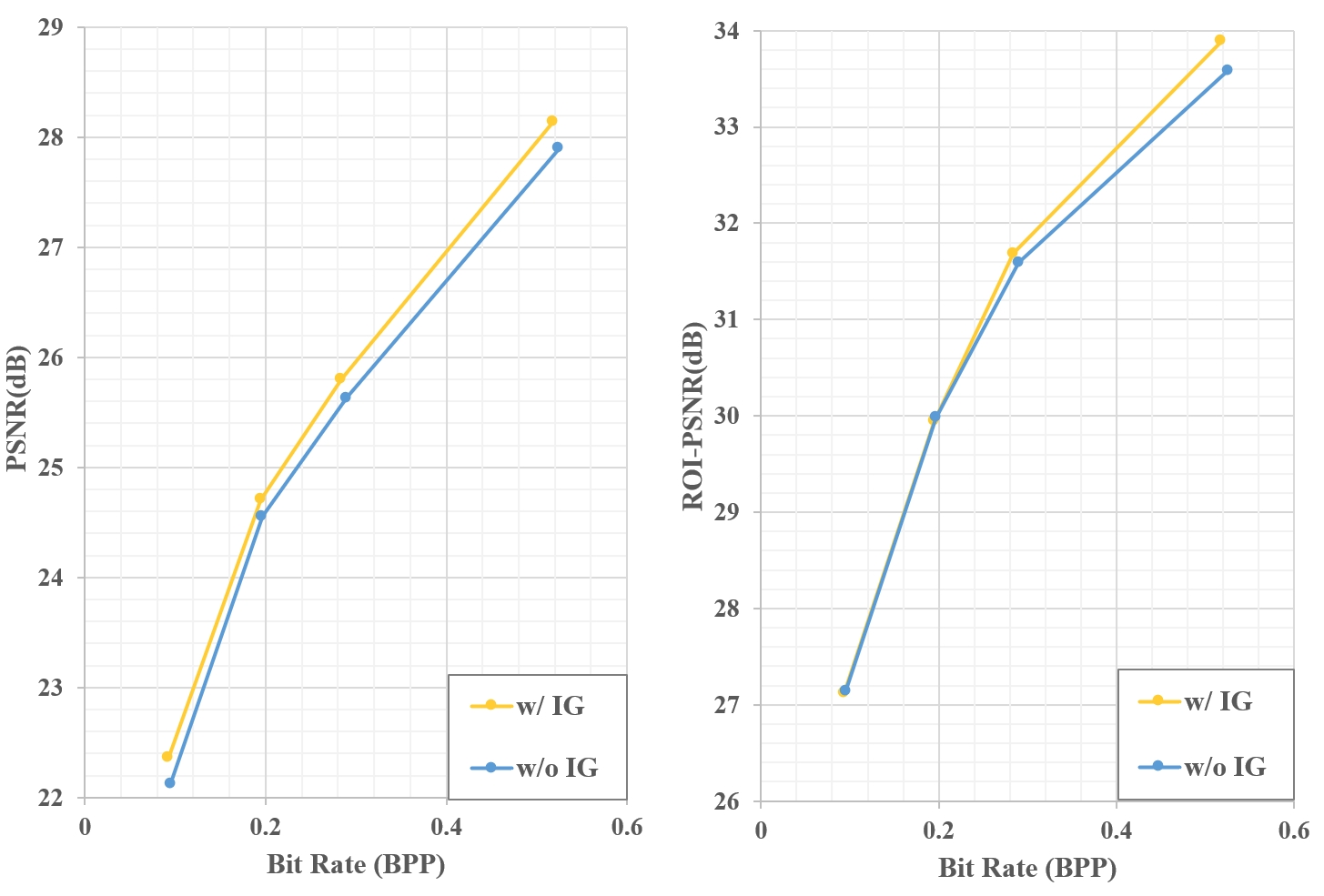}
\caption{RD performance of the IG ablation on the COCO validation dataset. }
\label{IG}
\end{figure}
\noindent{\textbf{IG ablation.}} To verify the effectiveness of the IG sub-module, we conduct an ablation study of the IG sub-module. Specifically, we perform our method with IG (w/ IG) and without IG (w/o IG), respectively. The ablation results are shown in Fig. \ref{COCO_dog}, where the reconstructed images with IG and without IG sub-module are shown in the second column (\emph{i.e.}, (b) and (e)) and the fourth column (\emph{i.e.}, (d) and (g)), respectively. It can be observed that the quality (PSNR) of the reconstructed image in (b) and (e) is better than that in (d) and (g) even under smaller BPP conditions. Besides, we also conduct the IG ablation across the entire COCO validation dataset and the experimental results are shown in Fig. \ref{IG}. It can be observed that both the PSNR and ROI-PSNR with the IG sub-module are higher than those without the IG sub-module. All these results demonstrate that the latent RDO prior generated with the IG sub-module benefits the latent representation of the source image. This also aligns with our claim on the effectiveness of the latent RDO prior. It should be mentioned that this is the first initial attempt that using the latent RDO prior to boosting the performance of the ROI-based deep image compression.

\begin{figure}[t]
\centering
\includegraphics[width=2.5in]{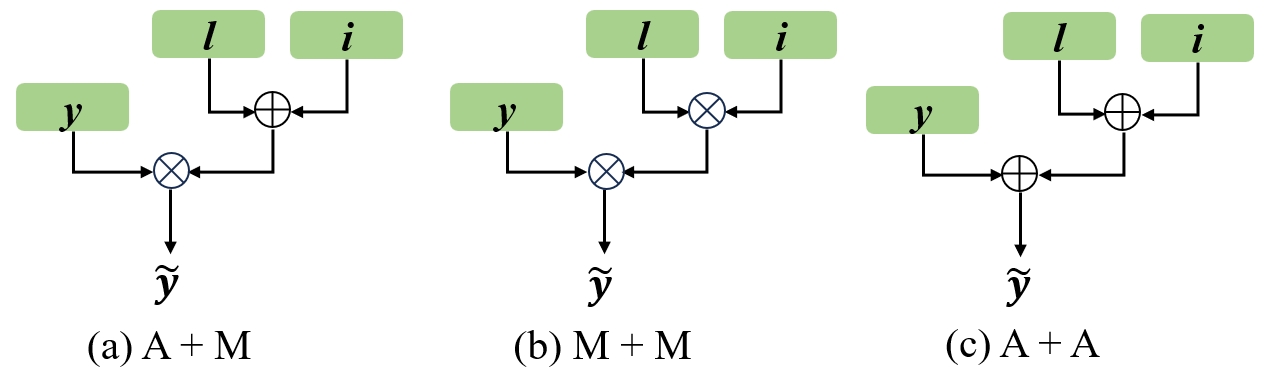}
\caption{The detailed architectures of different fusion methods. $\bigotimes$ and $\bigoplus$ denote the element-wise multiplication operation and the element-wise addition operation, respectively. }
\label{ablation}
\end{figure}

\begin{figure}[t]
\centering
\includegraphics[width=3in]{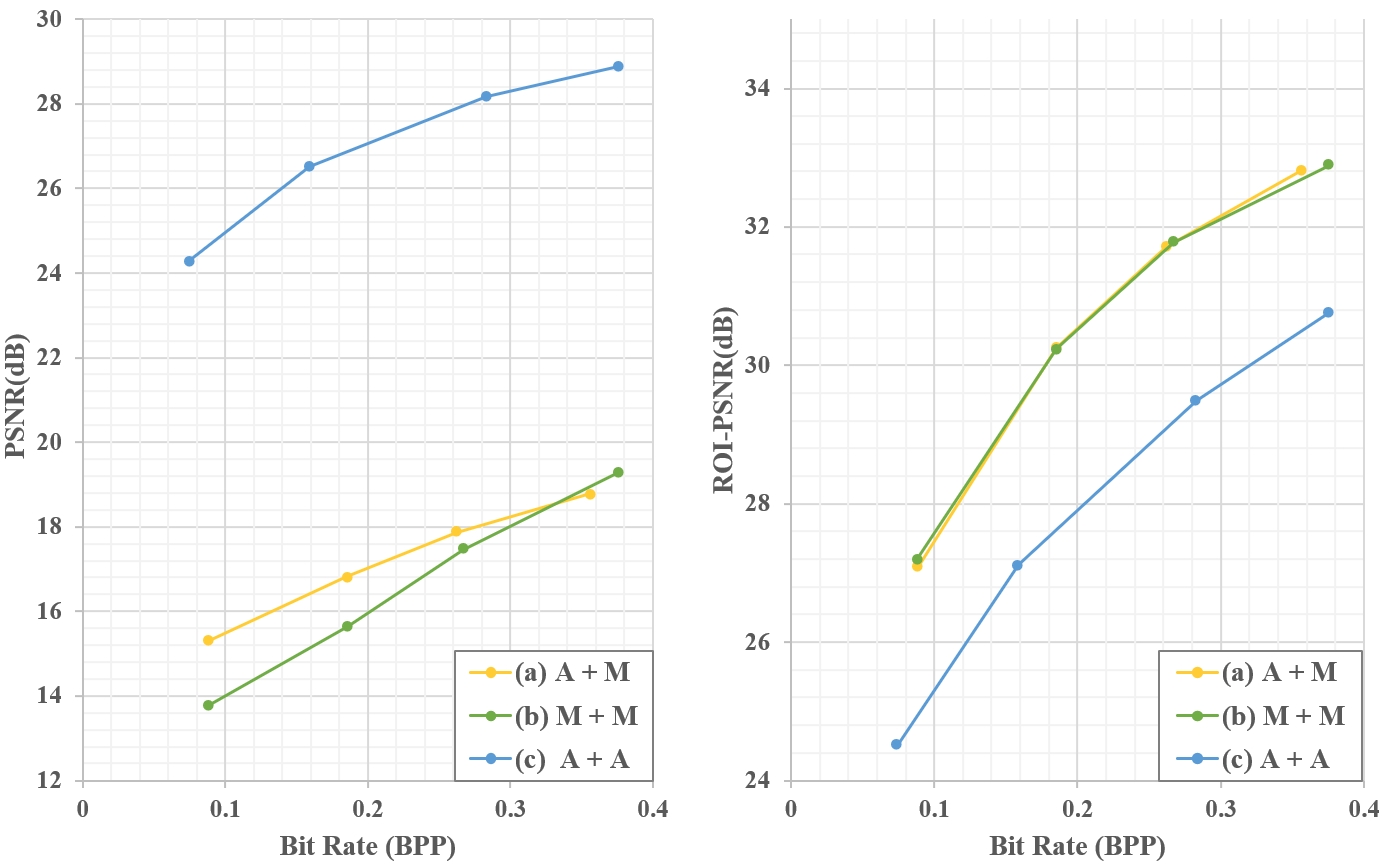}
\caption{RD performance of different fusion methods on COCO validation dataset. 
}
\label{fusion}
\end{figure}

\noindent{\textbf{Fusion method ablation.}}
To optimize the latent representation of the image, we fuse the latent spatial prior $l$, latent RDO prior $i$, and latent representation $y$ together. Specifically, we first apply an element-wise addition operation between $l$ and $i$ and then conduct an element-wise multiplication operation with $y$. We name this fusion process with Addition + Multiplication, termed A+M. To verify the effectiveness of the fusion method, we conduct the ablation study. Specifically, for comparisons, we ablating the A+M with two different settings. That is 1) apply element-wise multiplication operations of the three components directly, termed M+M and 2) apply element-wise addition of the three components directly, termed A+A. The detailed architectures are exhibited in Fig. \ref{ablation}. We conduct the ablation study on the COCO validation dataset and the results are shown in Fig. \ref{fusion}. It can be observed that our proposed A+M fusion method achieves outperformed results in terms of both PSNR and ROI-PSNR compared with the ablated methods. This verifies the effectiveness of our fusion method. 

After conducting all the ablation studies, we may conclude that three main components within the proposed method contribute to the compression efficiency gain, \emph{i.e.}, the MR sub-module, IG sub-module, and the fusion method. Specifically, first, we use the proposed MR sub-module to learn the latent spatial prior from the mask instead of the mask itself for guiding the latent representation of the source image, which provides a finer latent spatial prior of the ROI, especially for the ROI boundaries. This benefits both human perception and the downstream machine vision tasks. The MR ablation demonstrates our claims. Second, we use the IG sub-module to generate the latent RDO prior as another guide for the latent representation of the source image, which helps it understand which part of the latent representation contributes more to the source image in terms of RDO and avoids it learning from scratch. The IG ablation demonstrates our claims. Third, our proposed fusion strategy in the LMA module also offers distinct advantages in improving compression efficiency. The fusion method ablation demonstrates our claim as well.

\section{Conclusion}
In this paper, we have proposed a \emph{customizable ROI-based deep image compression} paradigm to address the various requirements of different users, especially for the mask acquisition as well as the reconstruction quality trade-off between ROI and non-ROI. To this end, we have proposed the Text-controlled Mask Acquisition (TMA) module that enables users to customize the ROI by inputting the semantic \emph{text}. Besides, we also have developed a Customizable Value Assign (CVA) mechanism which enables the reconstruction quality trade-off between ROI and non-ROI to be managed effectively. Moreover, unlike directly using the mask for guiding the latent representation of the image, we have developed a Latent Mask Attention (LMA) module that represents the mask as a latent spatial prior for guidance which is more helpful for achieving better Rate-Distortion (RD) performance, especially for the boundary regions of the ROI. Besides, to avoid the latent representation of the image learning from scratch, we also have fused the importance map as the latent RDO prior and the RD performance has been further improved. Extensive experiments on the Kodak and COCO datasets show that our proposed paradigm is feasible on the mask acquisition and has good enough zero-shot capabilities when comes across new semantic categories. Besides, the effectiveness of managing the reconstruction quality trade-off between ROI and non-ROI has also been demonstrated by assigning different values to the quality trade-off factor $\sigma$. Additionally, even by using the uniform mask as input, our paradigm can still achieve the SOTA results in both image reconstruction and machine vision tasks, such as object detection and instance segmentation.

Overall, this approach offers a paradigm for customizable ROI-based deep image compression to expand the functionality and applicability of ROI-based image applications in scenarios with diverse users and downstream tasks. It also paves a new way in the design of customizable image compression.

\vfill

\end{document}